\documentclass[sigconf]{acmart}

\AtBeginDocument{%
  \providecommand\BibTeX{{%
    \normalfont B\kern-0.5em{\scshape i\kern-0.25em b}\kern-0.8em\TeX}}}

\setcopyright{acmcopyright}
\copyrightyear{2018}
\acmYear{2018}
\acmDOI{10.1145/1122445.1122456}

\acmConference[Woodstock '18]{Woodstock '18: ACM Symposium on Neural
  Gaze Detection}{June 03--05, 2018}{Woodstock, NY}
\acmBooktitle{Woodstock '18: ACM Symposium on Neural Gaze Detection,
  June 03--05, 2018, Woodstock, NY}
\acmPrice{15.00}
\acmISBN{978-1-4503-XXXX-X/18/06}

\usepackage{subfig}
\usepackage{multirow}
\usepackage{caption}
\usepackage{dblfloatfix}

\newcommand{\textualpredictor}{textual predictor }
\newcommand{\visualpredictor}{visual predictor }
\newcommand{\Textualpredictor}{Textual predictor }

\setcopyright{rightsretained} 
\copyrightyear{2020}
\acmYear{2020}
\setcopyright{rightsretained}
\acmConference[KDD '20]{Proceedings of the 26th ACM SIGKDD Conference on Knowledge Discovery and Data Mining}{August 23--27, 2020}{Virtual Event, CA, USA}
\acmBooktitle{Proceedings of the 26th ACM SIGKDD Conference on Knowledge Discovery and Data Mining (KDD '20), August 23--27, 2020, Virtual Event, CA, USA}
\acmDOI{10.1145/3394486.3403381}
\acmISBN{978-1-4503-7998-4/20/08}

\begin{document}

\title{An Automatic Approach for Generating Rich, Linked Geo-Metadata from Historical Map Images}


\author{Zekun Li, Yao-Yi Chiang}
\affiliation{%
  \institution{University of Southern California}
  \city{Los Angeles}
  \state{CA}
  }
 \email{[zekunl, yaoyic]@usc.edu}

\author{Sasan Tavakkol}
\affiliation{%
  \institution{Google Research}
  \city{New York}
  \state{NY}
  }
 \email{tavakkol@google.com}
 
 \author{Basel Shbita}
\affiliation{%
  \institution{University of Southern California}
  \city{Los Angeles}
  \state{CA}
  }
 \email{shbita@usc.edu}
 
\author{Johannes H. Uhl, Stefan Leyk }
\affiliation{%
  \institution{University of Colorado Boulder}
  \city{Boulder}
  \state{CO}
  }
 \email{[johannes.uhl,stefan.leyk]@colorado.edu}

\author{Craig A. Knoblock}
\affiliation{%
  \institution{University of Southern California}
  \city{Los Angeles}
  \state{CA}
  }
 \email{knoblock@isi.edu}
 
\renewcommand{\shortauthors}{Zekun Li et al.}
\fancyhead{} 


\begin{abstract}
Historical maps contain detailed geographic information difficult to find elsewhere covering long-periods of time (e.g., 125 years for the historical topographic maps in the US). However, these maps typically exist as scanned images without searchable metadata. Existing approaches making historical maps searchable rely on tedious manual work (including crowd-sourcing) to generate the metadata (e.g., geolocations and keywords). Optical character recognition (OCR) software could alleviate the required manual work, but the recognition results are individual words instead of location phrases (e.g., ``Black'' and ``Mountain'' vs. ``Black Mountain''). This paper presents an end-to-end approach to address the real-world problem of finding and indexing historical map images. This approach automatically processes historical map images to extract their text content and generates a set of metadata that is linked to large external geospatial knowledge bases. The linked metadata in the RDF (Resource Description Framework) format support complex queries for finding and indexing historical maps, such as retrieving all historical maps covering mountain peaks higher than 1,000 meters in California. We have implemented the approach in a system called mapKurator. We have evaluated mapKurator using historical maps from several sources with various map styles, scales, and coverage. Our results show significant improvement over the state-of-the-art methods. The code has been made publicly available as modules of the Kartta Labs project at  https://github.com/kartta-labs/Project.
\end{abstract}



\maketitle

\section{Introduction}


%
Many professionally created historical maps are now accessible online as images through library repositories or map archives. These maps provide a unique opportunity for scientific studies that require long-term, historical geographic data, which do not exist elsewhere~\cite{chiang2020using}. However, these map images represent only a small fraction of the available map collections. For example, searching map records using WorldCat\footnote{WorldCat is a catalog system that indexes library collections from more than 100 countries and 17,900 libraries.} shows that less than 4\% of all map records in the US are digitized and published online. In contrast, the remaining 96\% of the map records are paper maps. With the advances in low-cost, high-speed automated scanners, the main reason that prevents these paper maps from being published online is the tedious manual process to compile metadata for the maps once they are scanned as images. At the University of Southern California, cataloging a map sheet requires about 30 minutes of manual work by a professional library curator, which would require 30 years to process all 125,000 historical topographic maps from the USGS (United States Geological Survey). As a result, even after paper maps are scanned into images, these map images are still not searchable because they do not come with any metadata, such as the geographic coordinates of the map center, the map scale, and place-related information on the maps.  For making all historical maps accessible, the first and crucial step is to increase the level of automation in generating a rich set of metadata for these maps so that they can be indexed and searchable by location and keywords. 

\begin{figure}[h]
\centering
\subfloat{\includegraphics[width=0.5\columnwidth,height=0.45\columnwidth]{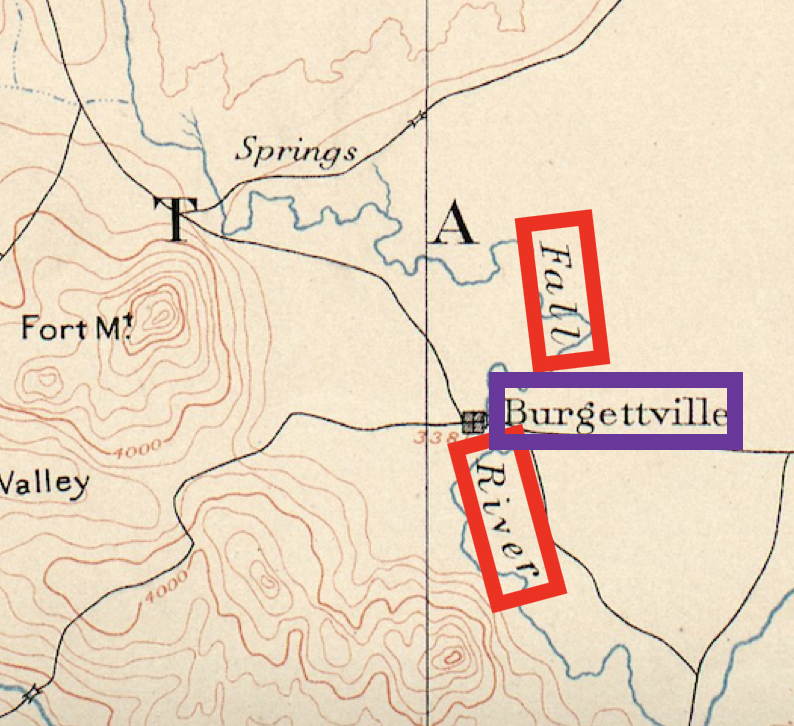}}
\subfloat{\includegraphics[width=0.5\columnwidth,height=0.45\columnwidth]{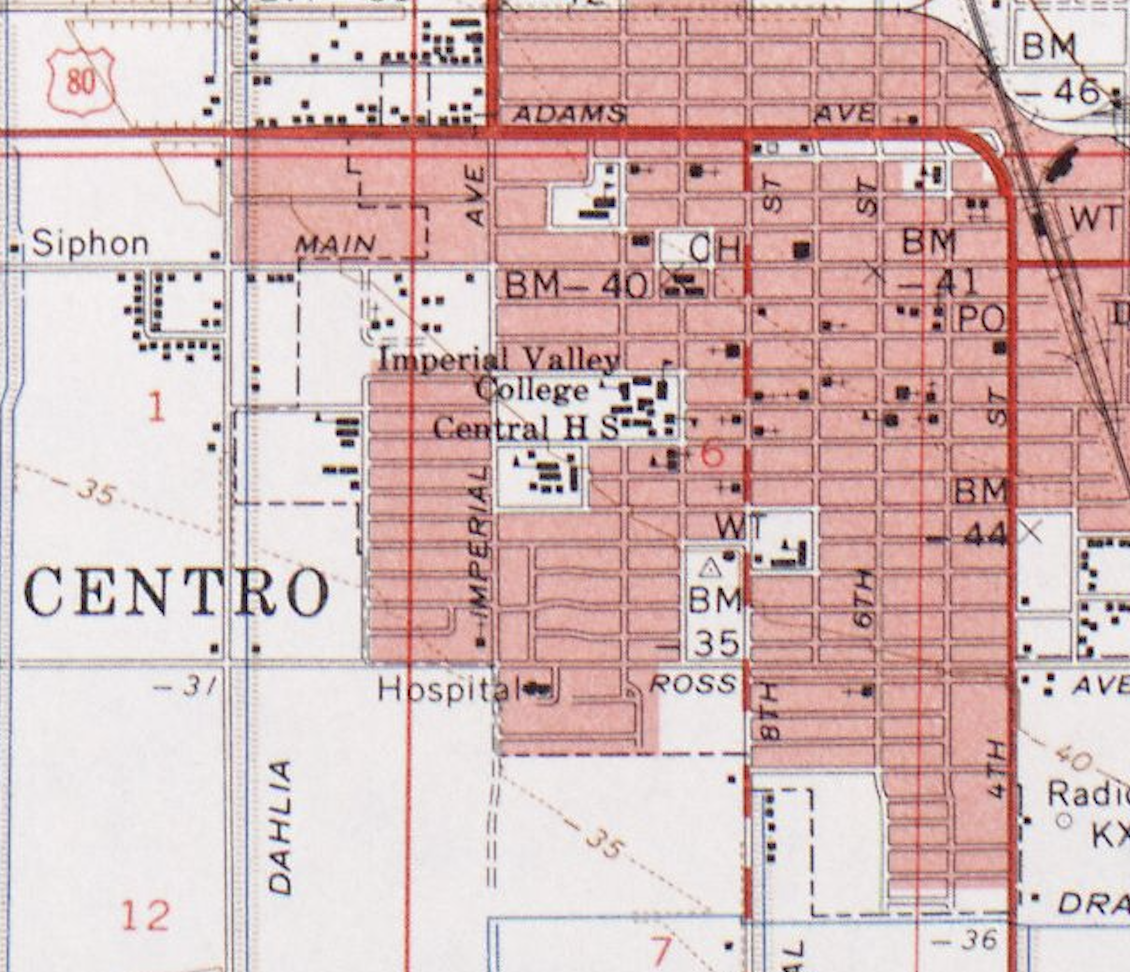}}
\caption{Challenges for text linking. Left: Distance is not the only deterministic factor for linking separate text regions. Right: Some maps contain densely-packed text regions.}
\label{fig:link_eg}
\end{figure}

Existing approaches to automatically generate map metadata rely on optical character recognition (OCR) tools that are designed for conventional document images but not maps~\cite{Chiang2015-jd}. For example, OCR tools focus on recognizing individual words and paragraphs in a document~\cite{yu2016recognizing, chiang2014survey, chiang2020using}. They do not handle map text that can have varying orientations and spacings. For example, each word of the place name ``Double Head Mountain'' on a map can be far away from each other and can have varying text orientations following the geographic features on the map. Also, on a map, text regions that are close to each other are not necessarily a part of the same location phrase (e.g., a place name). In Figure~\ref{fig:link_eg}, the word \textit{Fall} is close to the word \textit{Burgettville}, but they are not a part of the same phrase. The word \textit{Fall} should be linked to the word \textit{River} to constitute the location phrase \textit{Fall River}. Therefore the distance between two words is not sufficient to predict the linkage between individual words. Figure ~\ref{fig:link_eg} also shows a map area that contains densely packed text labels with the same font style and font size, which makes automated processes for generating map metadata challenging.

This paper presents a complete approach and its implementation, mapKurator, to automatically generate a rich set of metadata from map images using their text content. mapKurator first detects individual words from a map using an OCR tool (e.g., ~\cite{liao2018textboxes++,visionapi}) and then employs a deep neural network to link individual words in the OCR results (e.g., ``Los'' and ``Angeles'') to complete location phrases (e.g., ``Los Angeles'') based on their text and visual information (Section~\ref{sec:linking}). Next, mapKurator uses a geocoding service to find candidate geolocations for each detected location phrases and then and identifies spatial clusters of the candidates to remove unlikely geolocations and generate approximate geolocation for the map (Section~\ref{sec:geocoding}). Geocoding is the process of comparing text strings of addresses or place names to a large set of georeferenced data to generate the geocoordinates of the text strings (e.g., ~\cite{ge2005address, rashidian2018easergeocoder}). Once the approximate geolocation of the map is identified, mapKurator queries the LinkedGeoData to match the location phrases to entities in the LinkedGeoData using their location and text similarity (Section~\ref{sec:entity_matching}). The result is a set of metadata that is linked to large external geospatial knowledge bases (i.e., LinkedGeoData). The linked metadata in the RDF (Resource Description Framework) format support complex queries for finding and indexing historical maps, such as retrieving all historical maps covering mountain peaks higher than 1,000 meters in California.

\section{Overall Approach to Generate Map Metadata}
The overall approach of mapKurator for generating map metadata includes three major modules. In the first module, mapKurator generates complete location phrases from the input map. In the second module, mapKurator geolocates the place phrases and determines an approximate geolocation of the map. In the last module, mapKurator matches the georeferenced place phrases to entities on the LinkedGeoData to generate linked metadata for the input map.

\subsection{Generate Location Phrases from Words}~\label{sec:linking}
The goal for this module is to construct full location phrases from single words by first discovering the words that are in the same location phrase (text linking) and then sorting the words within each location phrase (e.g., ``Los'' should be before ``Angeles''). We formulate this text linking process as a query-retrieval problem. First, mapKurator runs text detection and recognition tools, such as Google Vision API ~\cite{visionapi}, to extract separate words and their bounding boxes. Then, mapKurator takes a detected text region as the query word and returns the potentially linked text regions by examining all other text regions on the map using its \textualpredictor and \visualpredictor. mapKurator repeats the querying process for every single text region on the map to construct a graph of the linked words. The \textbf{\textualpredictor}in mapKurator makes the first prediction to determine if a pair of words are linked together (i.e., in the same location phrase) using their textual information. \Textualpredictor is a binary classification model that looks at all the individual text regions on the map given a query region and filters out the ones that are unlikely to be linked to the query region. The purpose of the \textualpredictor is to effectively narrow down the search space for visual predictor. The \textbf{\visualpredictor}exploits the prediction from \textualpredictor and utilizes image information to refine the prediction. Visual predictor relies on a semantic segmentation model to harness the visual context and gives pixel-level prediction on the linkage relationship. Finally, the \textbf{consensus module} in mapKurator combines the output from both predictors and makes the final decision. By combining the results from the two models, mapKurator could achieve a better F1 score compared with each separate model. mapKurator uses the final linkage decision to construct a graph and find its connected components. mapKurator sorts the elements in each connected component according to a simple heuristic to constitute each location phrase.


\begin{figure*}[t]
\begin{center}
\includegraphics[width=\linewidth]{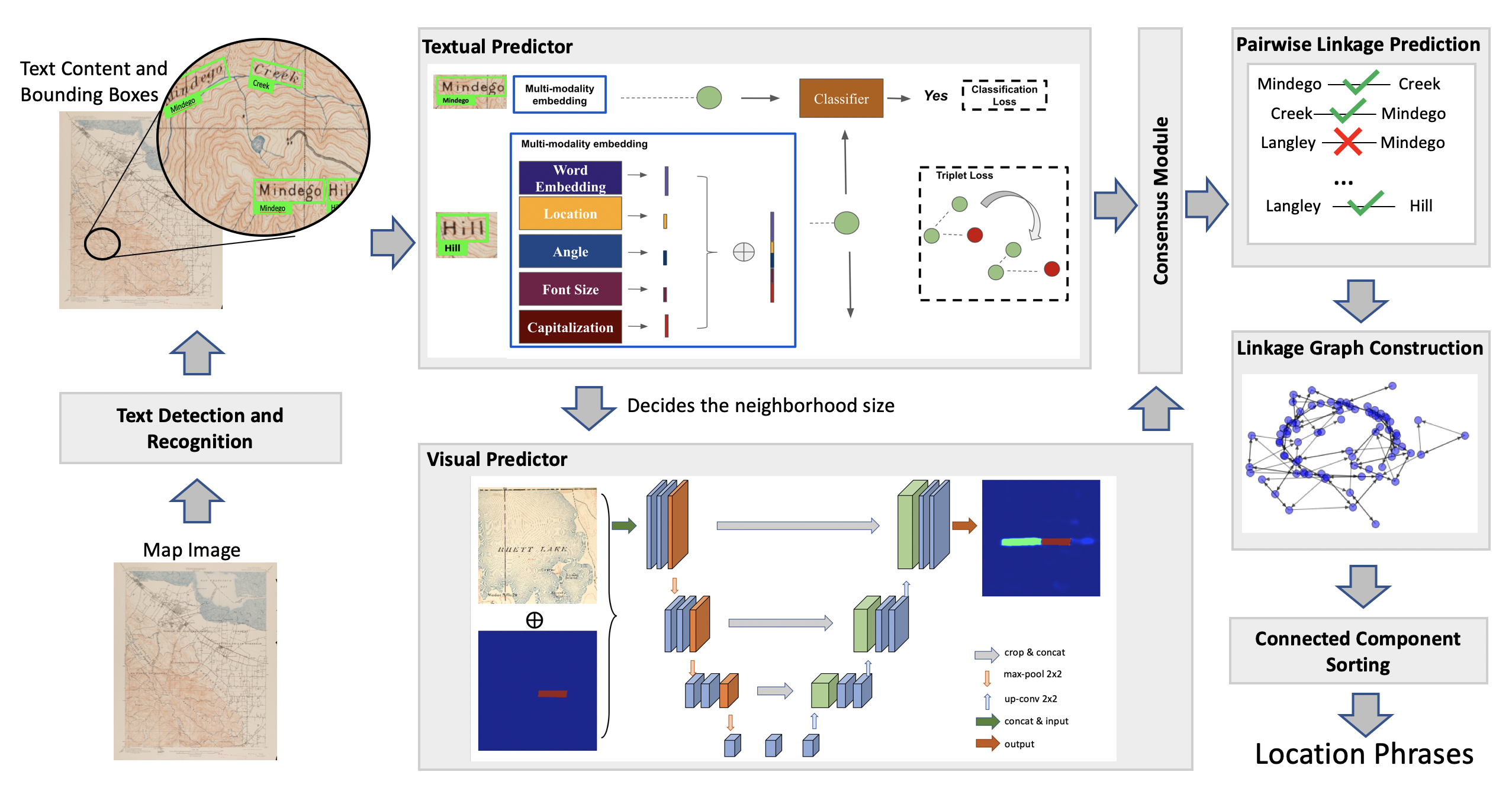}
\end{center}
   \caption{Generating location phrases from historical maps. Our method takes full map images as input, and runs text detection and recognition as a preprocessing step which yields recognized text content and predicted bounding boxes. Both \textualpredictor and \visualpredictor are working towards pairwise linkage prediction, whereas \textualpredictor considers text-related information, and \visualpredictor considers image-related information. Text-related information is the information that can be derived from the text region itself, such as location, orientation angle, word embedding, etc. While the \textbf{visual information} comes from the neighborhood of a specific text region on the image. After pairwise linkage prediction, we construct a graph, $G=\{N,E\}$, whose nodes, $N$, are separate text regions and edges, $E$, are predicted linkages. By computing the connected components and sorting their elements, we construct the location phrases. }
\label{fig:outline}
\end{figure*}


\subsubsection{Textual Predictor with Deep Metric Learning}

Given a pair of text regions such as ``Jefferson'' and ``Street'', we want the \textualpredictor module to predict whether they belong to the same location phrase. If they do belong to the same phrase, which is true for this case, the classifier should output 1, otherwise 0. We call the pair ``Jefferson'' and ``Street'' a \textbf{positive pair}, and a pair like ``Jefferson'' and ``Western'' that do not belong to the same location phrase, a \textbf{negative pair}.

The \textualpredictor in Figure ~\ref{fig:outline} takes pairs of text regions as input. The multi-modality embedding module in \textualpredictor utilizes the textual information associated with the input text regions. We define the inputs for text region $B_i$ as following: text content $t_i$; center location $c_{xi}$, $c_{yi}$; angle $a_i$ which is the clock-wise bounding box orientation angle with 0 degrees corresponding to a vertical and 90 degrees corresponding to a horizontal orientation; average font area size $f_i = b_i * h_i / len(t_i)$ where $b_i$ and $h_i$ are the width and height of the bounding box; capitalization $p_i$ which equals to 1 if \textbf{all} the characters in the word are in upper case, otherwise 0. Both location and font size are normalized to [-1,1] using formula $2x/(x_{max} - x_{min}) - 1$, and the angle is normalized to [0,1]. We use the Word2Vec ~\cite{mikolov2013efficient} model to embed the text content $t_i$ as a 50 dimensional feature vector $w_i$. The Word2Vec model weights are initialized by the pre-trained weights on the Wikipedia 2014 + Gigaword 5th Edition corpora (6B tokens, 400K vocab) using the GloVe ~\cite{jeffreypennington2014glove} algorithm. Then we concatenate $\{w_i, c_{xi},c_{yi},a_i, f_i, p_i\}$ to obtain the textual feature representation $r_i$ of text region $B_i$. To determine if the two text regions should be connected together, we perform binary classification on the textual features. Also, we enforce the textual feature to be similar in embedding space if two regions are connected. We use binary cross-entropy as the classification loss:
\begin{equation}
    L_{CE} = -\frac{1}{N}\sum_{i=1}^{N}y_i log(p_i) + (1-y_i)log(1-p_i)
\end{equation}
\noindent where $y_i$ is the ground truth label for sample $i$, $p_i$ is the prediction for sample $i$, and $N$ is the number of samples. We also apply triplet loss ~\cite{taigman2014deepface, hoffer2015deep}on the embedding space to enforce similar features to be close to each other, and dissimilar features far away. The loss equation reads as:

\begin{equation}
    L_{tri} = \sum_{i}^{N}[||r_i^a - r_j^p||^2_2 - ||r_i^a - r_k^n||^2_2 + \alpha]_+
\end{equation}
where $r_i = concat(w_i, c_{xi},c_{yi},a_i, f_i, p_i)$ and $r^a$ is the anchor feature in triplet loss, $r^p$ is the positive feature, and $r^n$ is the negative feature.

\subsubsection{Visual Predictor with Weekly Supervised Segmentation}
The \visualpredictor first draws a minimum bounding box that includes all the retrieved positive candidates from the \textualpredictor. Then it predicts the probability of a pixel belonging to a text region that should be connected to the query text region. The bounding box is expected to be relatively small compared with the full map image since the \textualpredictor has removed the majority of the negative samples. Thus the previous step saves us from looking at the whole image so that we could focus on a smaller neighborhood of the query text region. 

\begin{figure}[h]
\captionsetup[subfloat]{farskip=2pt,captionskip=2pt}
\centering
\subfloat{\includegraphics[width=0.99\columnwidth]{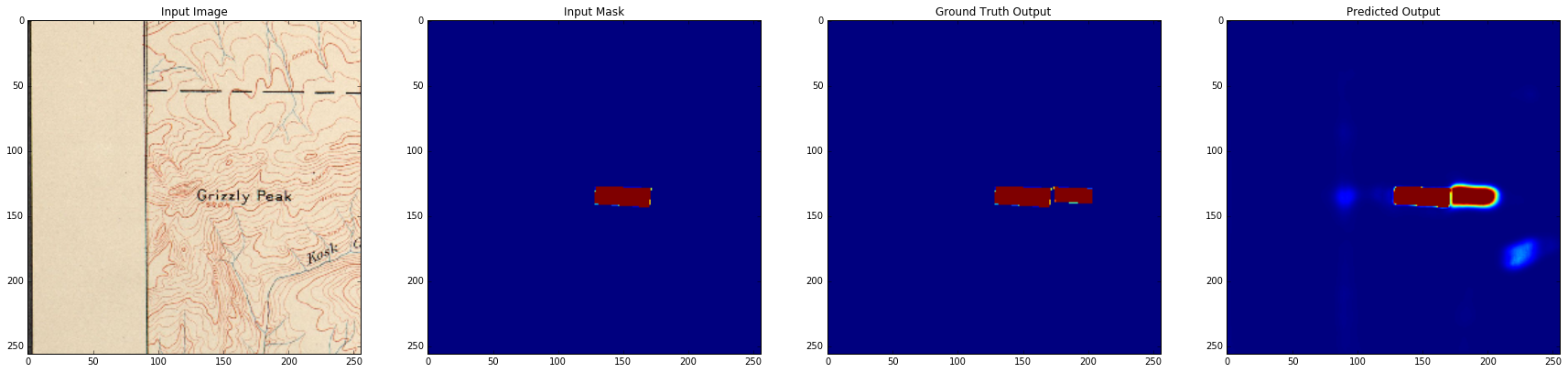}} \\
\subfloat{\includegraphics[width=0.99\columnwidth]{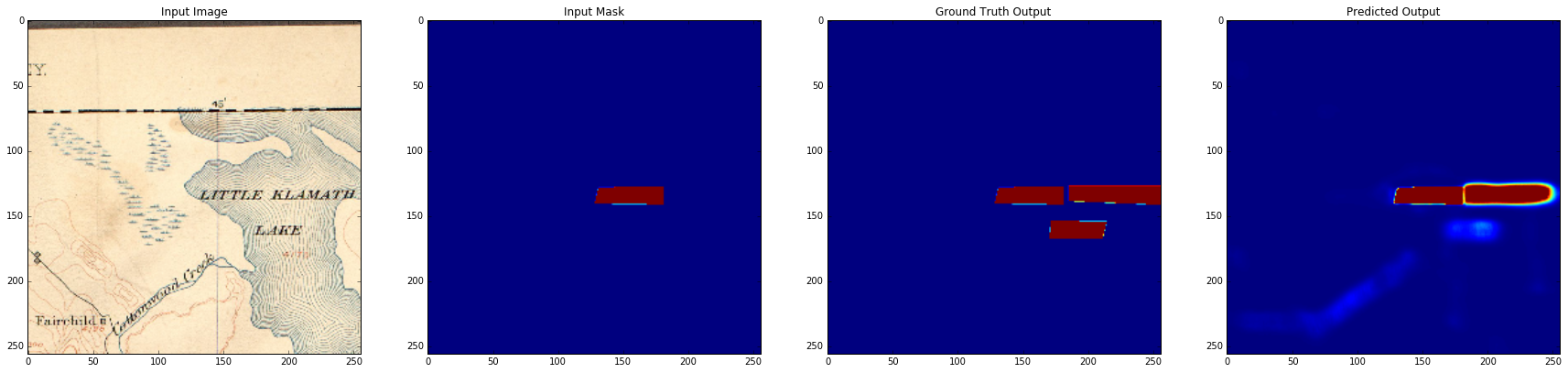}} \\
\subfloat{\includegraphics[width=0.99\columnwidth]{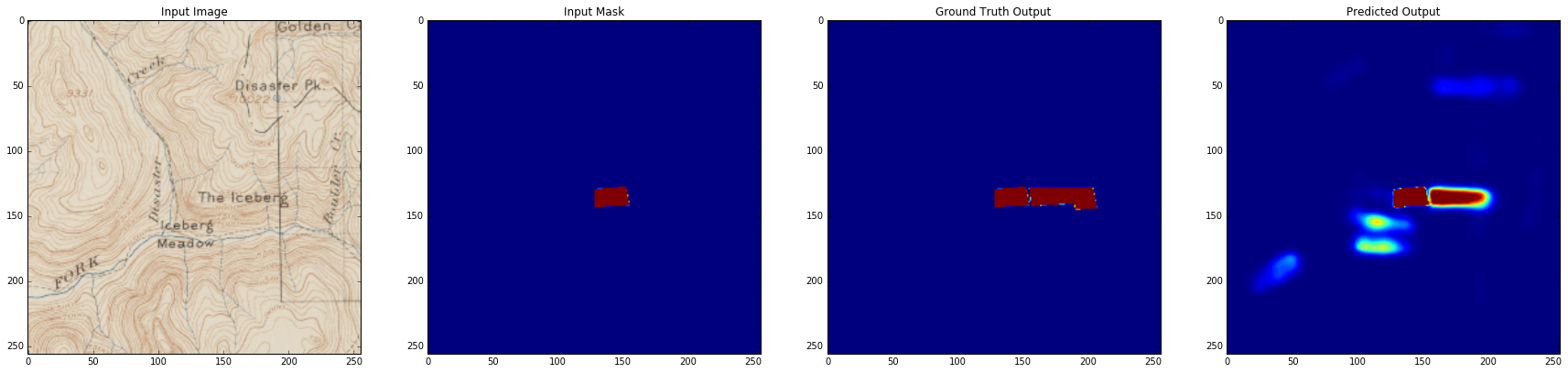}} \\
\caption{Visualization of the input, ground-truth and output of weakly supervised segmentation module with U-Net structure. The first two columns are inputs, the third column is the ground-truth output, and the last column is the prediction of the model.}
\label{fig:unet-output}
\end{figure}

\begin{figure}[t]
\begin{center}
\includegraphics[width=\linewidth]{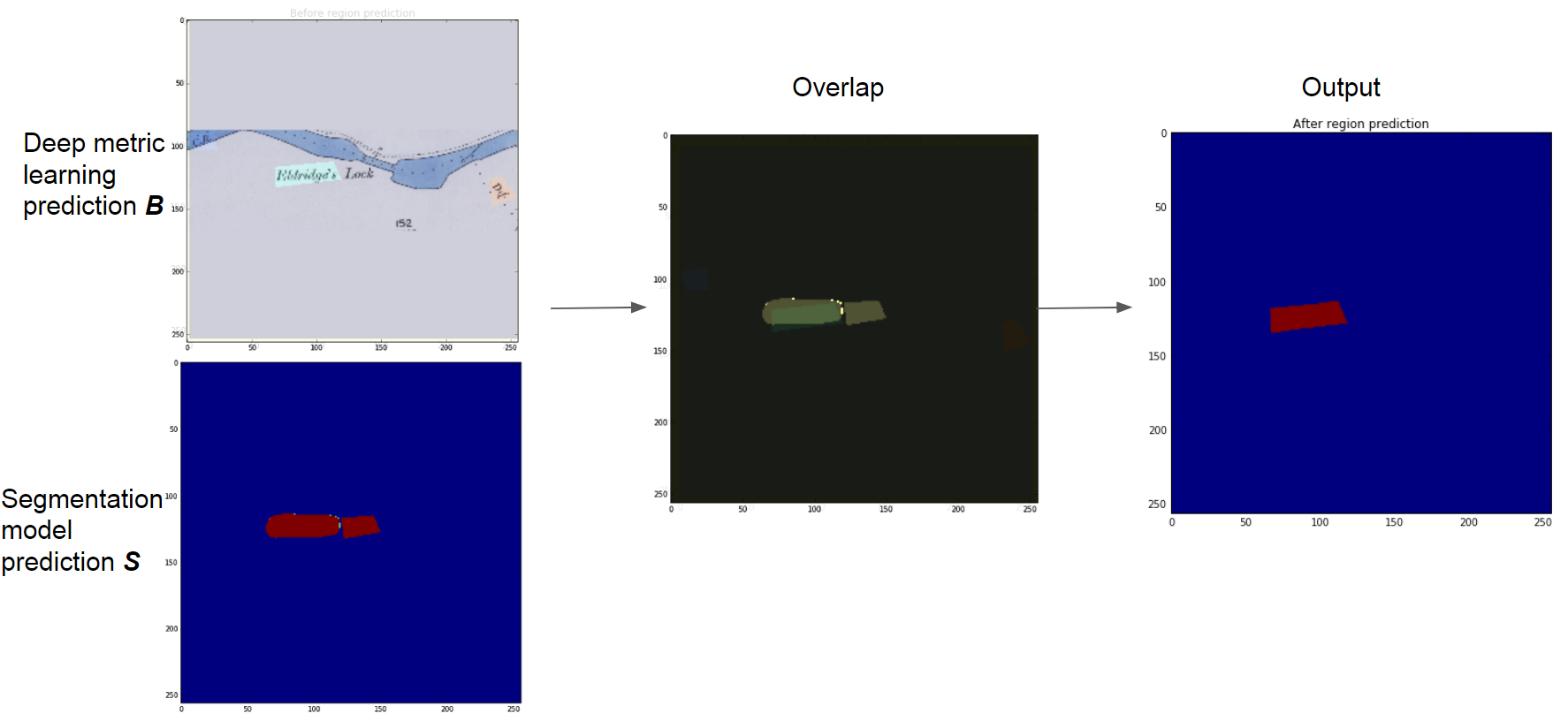}
\end{center}
   \caption{Consensus module to find the common retrievals of \visualpredictor and \textualpredictor }
\label{fig:combiner}
\end{figure}

We use the U-Net structure ~\cite{ronneberger2015u} for our weakly supervised segmentation model. The model takes two inputs: (A) cropped RGB image around the neighborhood of the query text region (see the first column of Figure ~\ref{fig:unet-output}) and (B) a mask that highlights the query text region (see the second column of Figure ~\ref{fig:unet-output}). The two inputs have the same size, letting us concatenate the two inputs along the channel dimension to form a 4-channel tensor as the direct input for the model. The output of the model is a probability map that indicates the probability of a pixel belonging to a connected text region of the query region. From the probability map (shown in the last column in Figure ~\ref{fig:unet-output}, we use the threshold $p=0.5$ to binarize the prediction to produce the second input (see lower left image in Figure \ref{fig:combiner}) for consensus model.  We use \texttt{sigmoid} activation for the last layer of U-Net and binary cross-entropy for the loss.

\subsubsection{Consensus Model}
We combine the results from \textualpredictor and \visualpredictor using a consensus module. Assume that the \textualpredictor produces $M$ candidate text regions for the query region $B_q$, and each candidate text region is referred to as $B_i$ with $i\in[1,M]$. Assume that the minimum bounding box that contains candidate text regions is of size $P\times Q$, we pad the bounding box to a square whose edge length equals to $max(P,Q)$. The padding color is ($m_r$,$m_g$, $m_b$) where $m_r$, $m_g$, $m_b$ define the mean color density for each channel. To facilitate batch training, we resize the padded image of size $max(P,Q)\times max(P,Q)$ to size $N \times N$ with either upsampling or downsampling and feed into the \visualpredictor. Thus the output for \visualpredictor $S$ is a tensor of size $1\times N\times N$ where $N$ = 256 in our experiments. We use the following metric to determine if $B_i$ is a consensus for both modules. 
\begin{equation}
    \frac{1}{||B_i||}(\sum_j^N\sum_k^N(B_i \otimes S_{jk})) > \theta
\end{equation}
where $\otimes$ is the element-wise multiplication, and $\theta$ is a hyper-parameter that serves as a threshold. This metric checks if the average probability of the overlapping region from textual predictor and visual predictor is higher than a certain threshold. We use $\theta = 0.5$ as the threshold in the experiments.

\subsubsection{From pairwise linkage prediction to location phrases} \label{sec:pair2final}

After we obtain pairwise predictions, we construct a directed graph $G = \{N,E\}$ where nodes $N$ are the separate text regions from the map, edges $E$ are the predicted linkages from the previous step. If the region $r_j$ is classified as a linked region with query region $r_i$, we draw a link from node $i$ to node $j$ on the graph. We can use either \textit{strongly} connected components (SCC) or \textit{weakly} connected components(WCC) to find the clusters where each cluster should contain all the text regions that form a full location phrase. A strongly connected component is identified if there is a path in \textit{each} direction between each pair of vertices inside the component. A weakly connected component is identified if each node can either reach another node or is reachable from another node. The type of errors for WCC and SCC are different, as that WCC tends to have "adding words" error and scc tends to have "missing words" error. In the experiment section, we use strongly connected component to find the clusters.


The output of connected components are groups of words that can form location phrases. To generate the final location phrase, we need to sort the words in the same set and put them in a sequence. We use a simple method that works well for most of the cases: sort ascendingly for the x-coordinates of the bounding box centers. Other methods can be developed to improve the sorting, and we leave that as future work. 

\subsection{Detect Geo-locations of Location Phrases}\label{sec:geocoding}

\begin{figure*}[h]
  \centering
  \includegraphics[width=\linewidth]{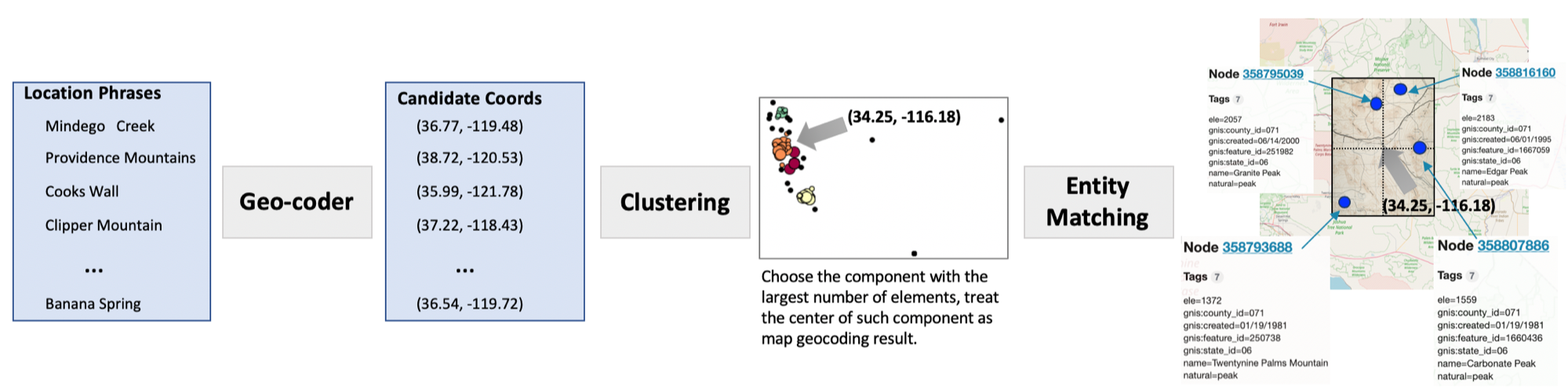}
  \caption{Pipeline for geolocalization and entity-matching. The input can either be location phrases or single words extracted from map images. This figure uses the location phrase as an example, but the steps are the same with single words. Based on different inputs, we call the process word-by-word or phrase-by-phrase geolocalization. The output for phrase-by-phrase geocoding is a list of candidate coordinates. We use DB-SCAN to cluster the coordinates and choose the center of the cluster that has the largest number of elements as the geolocalization output. Given the geo-coordinate of the cluster center, we perform entity matching that associates the map with external databases such as GeoLinkedData.  }
  \label{fig:geocoding}
\end{figure*}

\begin{figure}[h]
  \centering
  \includegraphics[width=\linewidth]{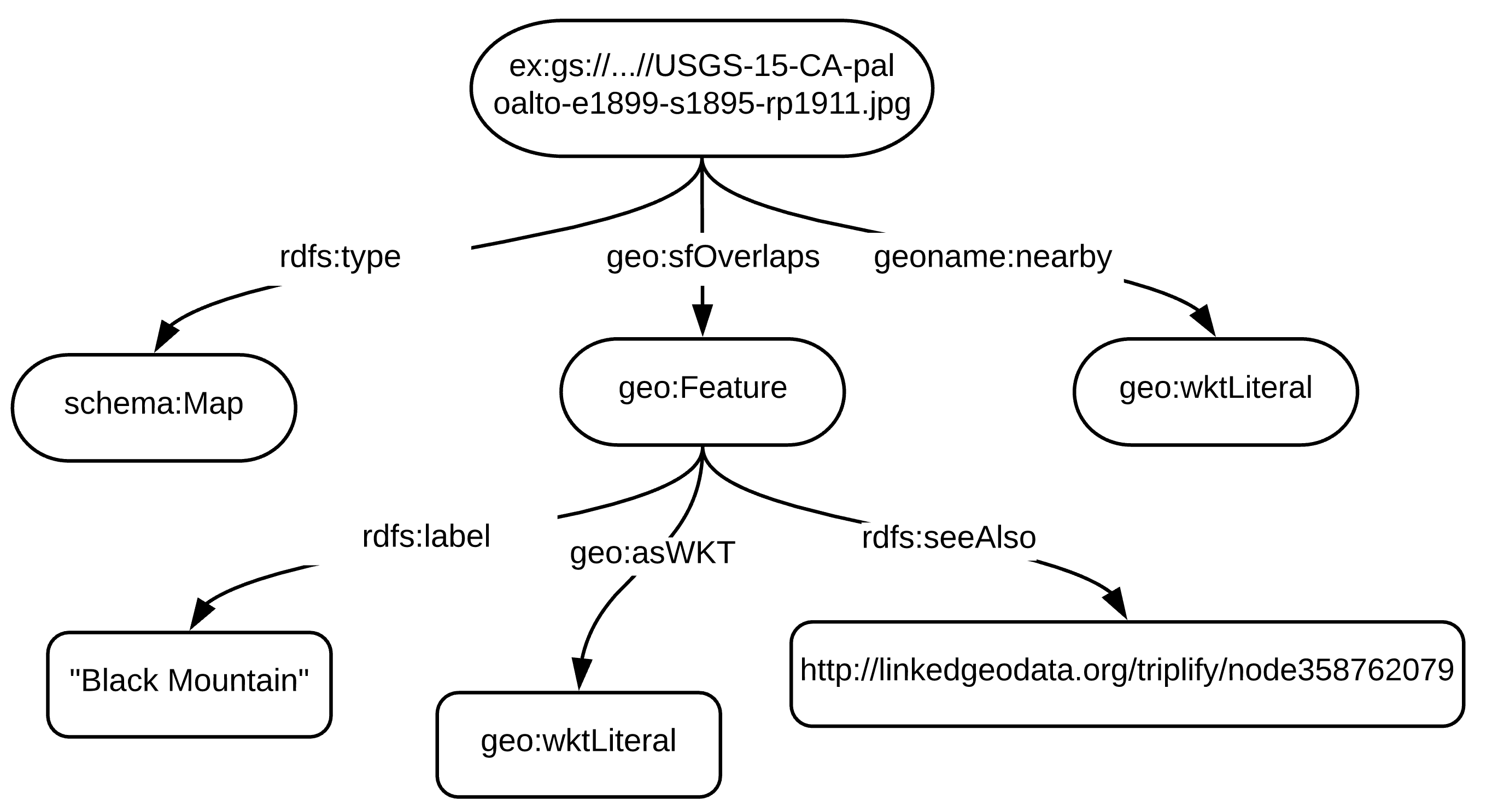}
  \caption{RDF schema for linked-historical maps. We convert each raster map image into a node in the linked graph, and each node has three properties "type", "sfOverlaps" and "nearby". "sfOverlaps" describes the text regions that are in the map and "nearby" describes the rough location of the map study area. }
  \label{fig:rdf}
\end{figure}

Tavakkol et al. ~\cite{tavakkol2019kartta} introduce a baseline model that concatenates all detected words from a map image into a paragraph, and send the paragraph to Google Geocoding API to produce the geo-coordinates. Although Google Geocoding API works on contemporary data instead of historical data directly, it still carries some advantage: it is a scalable publicly available service and it accounts for minor misspellings and discrepancies. One reason they construct a single paragraph out of the detected words is that they do not have a model to generate phrases out of the individual words. As mentioned earlier, individual words often are not meaningful per se. Furthermore, by calling the Geocoding API only once and choosing the first candidate, they eliminate the need for an algorithm to rank the candidate locations. We improve their framework by generating the location phrases, sending them individually to the Geocoding API, and utilizing a clustering algorithm to produce the geolocation of a map given many candidates that are returned from multiple calls to the Geocoding API. Note that the Geocoding API might also return multiple candidates for a single call. The clustering method we use is DBSCAN~\cite{ester1996density}, which locates the regions that have high density of geolocalized coordinates returned by the Geocoding API. We use the centroid of the cluster that contains the largest number of elements as the final geocoordinates for the map.  The workflow for geocoding is shown in Figure ~\ref{fig:geocoding}. For the sake of comparison, we once send the texts phrase-by-phrase and once word-by-word to the Geocoding API. For example, in the phrase-by-phrase method, the Geocoder takes ``Vermont Street'' as input while in word-by-word method, the Geocoder takes ``Vermont'' and ``Street'' separately and returns two groups of candidates. In the experiments, we compare our phrase-by-phrase geocoding and word-by-word geocoding with the baseline. 

\subsection{Matching Entities in Historical Maps to OpenStreetMap} \label{sec:entity_matching}
In the last few decades, researchers have made efforts to structure the contemporary map data into a linked form and have built databases such as LinkedGeoData. Such databases allows queries on contemporary map databases such as OpenStreetMap. However, historical maps are inherently unstructured images. Thus geospatial queries are difficult to run on this type of data. In this paper, we use common vocabularies and the RDF format to represent the automatically generated metadata from historical maps and link the metadata to the existing LinkedGeoData database using the inferred geographical coordinates obtained from Section~\ref{sec:geocoding}. Figure ~\ref{fig:rdf} shows the common vocabularies and the ontology in mapKurator. Section ~\ref{sec:case_study} describes details of the namespaces in the ontology.  This schema provides information for the map image type, coordinates of the map center, location names that are in the map image, and their associated information from other sources (e.g., GeoLinkedData).

The ``sfOverlaps" predicate connects a map subject with a text region object, denoting that the location name is contained in the map image. The  ``seeAlso" property of ``Feature" provides a URI of the location name in another database. When jointly queried with the external database, this property provides a way to backtrack to our Linked-historical-maps dataset. For example, if we are interested in finding maps that contain mountain(s), and LinkedGeoData provides us a set of geo-features with type mountain. We could easily know which ones of them are also contained in the Linked-historical-maps by comparing the URI, and further backtracking to the map image.

\section{Experiments}
In this section, we show the experimental results of mapKurator using several map sources. Our system is evaluated from two aspects: location phrase prediction and map geolocalization. We also provide a case-study of the linked map metadata. This section first describes the datasets used for the experiments, then elaborates on experiment details. 

\subsection{Datasets}

\textbf{United States Geological Survey (USGS) topographic maps}~\cite{usgs}: This dataset consists of 15 maps produced by the United States Geological Survey that covers the California State. All the text regions have been manually annotated and transcribed. There are about 293 text regions per map and 4,375 text regions in total. The ground truth geolocation is also provided. This dataset can be used to evaluate both location phrase prediction and map geolocalization tasks.  \\

\noindent\textbf{United Kingdom Ordnance Survey Maps} \footnote{Reproduced with the permission of the National Library of Scotland}~\cite{odmaps} consists of 10 maps published by Ordnance Survey and related bodies, including the War Office (ca. 1840s-1960s). These maps cover some of the regions in the United Kingdom. All the text regions have been manually annotated and transcribed. There are 216 text regions per map on average and 2197 text regions in total. The ground truth geolocation is \textbf{not} provided. Thus this dataset can only be used to evaluate location phrase prediction tasks. \\

\noindent\textbf{NYPL maps}~\cite{nypl}  consists of 500 maps collected by the New York Public Library that Tavakkol et al. ~\cite{tavakkol2019kartta} used to test their geolocalization algorithm. These maps are in the New York region. The text regions are \textbf{not} annotated or transcribed, but the ground truth geolocation is provided. Thus this dataset can only be used to evaluate the map geolocalization task. 

\begin{table*}
\centering
  \begin{tabular}{c|c|ccc|ccc}
    \toprule
    \multirow{2}{*}{Dataset } & \multirow{2}{*}{Map Name } &
      \multicolumn{3}{c}{\textbf{Textual Info. Only}} &
      \multicolumn{3}{c}{\textbf{Visual and Textual}}  \\
      & & {Prec.} & {Rec.} & {F1 } & {Prec.} & {Rec.} & {F1 }  \\
      \midrule
    \multirow{3}{*}{USGS } &60-CA-amboy-e1942 & 37.23 & 100.00 & \textbf{54.26} & 93.29 & 56.71 &\textbf{70.54} \\
    & 60-CA-amboy-e1943-rv1943 & 27.01 & 100.00 & \textbf{42.53} & 91.30 & 59.42 & \textbf{71.99} \\
    & 60-CA-modoclavabed-e1886 & 29.03 & 90.55 &\textbf{43.97} & 94.48 & 37.85 &\textbf{54.04}\\
    \midrule
    \midrule
     \multirow{3}{*}{OD } & 103681073 & 50.12 & 97.14 &\textbf{66.12} & 74.52  & 91.43 &\textbf{82.12}\\
    & 103681079 & 50.97 & 97.67 &\textbf{66.98} & 84.88 & 91.19 &\textbf{89.29} \\
    & 103681115 & 62.03 & 94.94 & \textbf{75.03} & 87.34 & 79.75 &\textbf{83.37} \\
    \bottomrule
  \end{tabular}
  \caption{Linkage prediction result on USGS and Ordnance Survey dataset. We use three images from each test set to compare the results of using only textual information versus using both textual and visual information. }
  \label{tab:linkage_prediction}
\end{table*}

\begin{table*}
\centering
  \begin{tabular}{c|c|ccc|ccc|c}
    \toprule
    \multirow{2}{*}{Dataset } & \multirow{2}{*}{Map Name } &
      \multicolumn{3}{c}{\textbf{Duplicate Phrases}} &
      \multicolumn{3}{c}{\textbf{Distinct Phrases}}  & \textbf{\# GT } \\
      & & {Prec.} & {Rec.} & {F1 } & {Prec.} & {Rec.} & {F1 }  &  \textbf{phrase}\\
      \midrule
    \multirow{3}{*}{USGS } &60-CA-amboy-e1942 & 44.72 & 63.16 & \textbf{52.36} & 54.64 & 57.61 &\textbf{56.08} & 114  \\
    & 60-CA-amboy-e1943-rv1943 & 52.45 & 69.03 &\textbf{59.61} & 63.43 & 65.38 & \textbf{64.39} & 155\\
    & 60-CA-modoclavabed-e1886 & 31.28 & 51.85 &\textbf{39.02} & 33.33 & 44.71 &\textbf{38.19}& 108\\
    \midrule
    \midrule
     \multirow{3}{*}{OD } & 103681073 & 74.54 & 85.41 &\textbf{79.61} & 62.50 & 71.42 &\textbf{66.67} & 48 \\
    & 103681079 & 84.00 & 91.30 &\textbf{87.50} & 73.91 & 82.92 &\textbf{78.16} & 69 \\
    & 103681115 & 58.33 & 68.29 & \textbf{62.91} & 55.31 & 63.41 &\textbf{59.08} & 41\\
    \bottomrule
  \end{tabular}
  \caption{Phrase prediction result on USGS and Ordnance Survey dataset. Note that the predicted phrases with the exactly same \textit{content and order} as the ground truth will be considered as a correct prediction. \textit{Duplicate Phrases} means the evaluation considers duplicate phrases in the maps as well. (e.g: "Modoc Lava Beds" may appear twice in the map and we evaluate both cases.) \textit{Distinct Phrases} means we remove the duplicate phraes for evaluation }
  \label{tab:phrase_prediction}
\end{table*}

\subsection{Experimental Result} \label{sec:result}
The main result from mapKurator is the location phrases and the map geolocation, which together, can generate a rich set of linked metadata. We evaluate the performance of mapKurator in generating each result component. For the location phrase generation, we evaluate mapKurator at two stages. The first stage is after pairwise prediction has been made, and the second stage is when the location phrase generation is complete. We show our result on the \textbf{Ordnance Survey} and \textbf{USGS} dataset since these two datasets have manually annotated bounding boxes, transcribed words and phrases. These two datasets have provided the bounding boxes and text content for individual text regions, thus we skip the text detection step and use these information directly as the input for linkage prediction model. Table~\ref{tab:linkage_prediction} shows the linkage prediction performance on these two datasets. In this table, we compare the result of using only textual information and the result of using textual and visual information. The precision is defined as $p_l = TP_l / (TP_l + FP_l)$, recall is defined as $r_l = TP_l / (TP_l + FN_l)$ and $F1 = 2p_l \cdot r_l/(p_l + r_l)$, where $TP_l$ is the number of predicted linkages that have been correctly classified, $FP_l$ is the number of predicted linkages that have been incorrectly classified, and $FN_l$ is the number of linkages that have been missed from prediction. We can see that $F1$ score is improved by a large margin when visual information is added. We also notice that when using only textual information, the recall is high, but the precision is low. This may be due to the fact that if only textual information is considered, the model lacks information about context the text regions belong to, which is similar to predicting the linkage relation on a blank background image. Thus, the network predicts relatively more positive linkages than the ground truth and incurs a high false-positive rate. By adding visual information, most of the false positives can be removed. Table~\ref{tab:phrase_prediction} shows the performance for full location phrase generation. The results are generated on the map-wise level by calculating how many location phrases have been correctly recognized. 
We show the numerical results for two cases: 1) For \textit{Duplicate Phrase}, we include duplicate phrases in evaluation 2) For \textit{Distinct Phrases}, we evaluates the distinct phrase correctness. For example, if the ground truth contains two "Modoc Lava Beds" and one "Black Crater", the prediction extracts two "Modoc Lava Beds" succesfuly but no "Black Crater". In the first metric, the recall is 2/3 and in the second metric the recall is 1/2.

We evaluated the performance on geolocalization by calculating the distance deviated from ground truth for the predicted latitude and longitude coordinates. In table ~\ref{tab:usgs_geolocalization_prediction}, We compare our \texttt{word-by-word} and \texttt{phrase-by-phrase} geocoding method described in Section ~\ref{sec:geocoding} with the \texttt{word-to-paragraph} method introduced in ~\cite{tavakkol2019kartta} on USGS dataset.  We show the error in both \texttt{kilometers} and in \texttt{scale}. The error metrics are defined bellow:

\begin{align}
    & Err_{km}(\textbf{g},\textbf{p}) = Haversine(\textbf{g},\textbf{p}) = \sum_i 2\arcsin( \sqrt{h_i}) \\
    & h_i = \sin^2(\frac{g^{lat}_i - p^{lat}_i}{2}) + \cos(g^{lat}_i)\cos(^p{lat}_i)\sin^2(\frac{g^{lng}_i - p^{lng}_i}{2})
\end{align}
where $\textbf{g}$ is the ground truth location and $\textbf{p}$ is the prediction.
The reason to use \textit{Haversine} distance instead of \textit{Euclidean} distance is that the Haversine distance measures the great-circle distance on a sphere, which is more appropriate in our setting while  Euclidean distance measures the straight-line distance between two points.
\begin{align}
    & Err_{scale}(\textbf{g},\textbf{p}) = \frac{Err_{km}}{Haversine(t_{min},t_{max})} 
\end{align}
where $t_{min}$ is the point at the corner of the map plot area that has the smallest latitude and longitude value, similarly, $t_{max}$ is the point with the largest value. Thus if $Err_{scale}$ is larger than 1, the map image does not cover the predicted geolocalization coordinates.

\begin{figure}[t]
\begin{center}
\includegraphics[width=\linewidth]{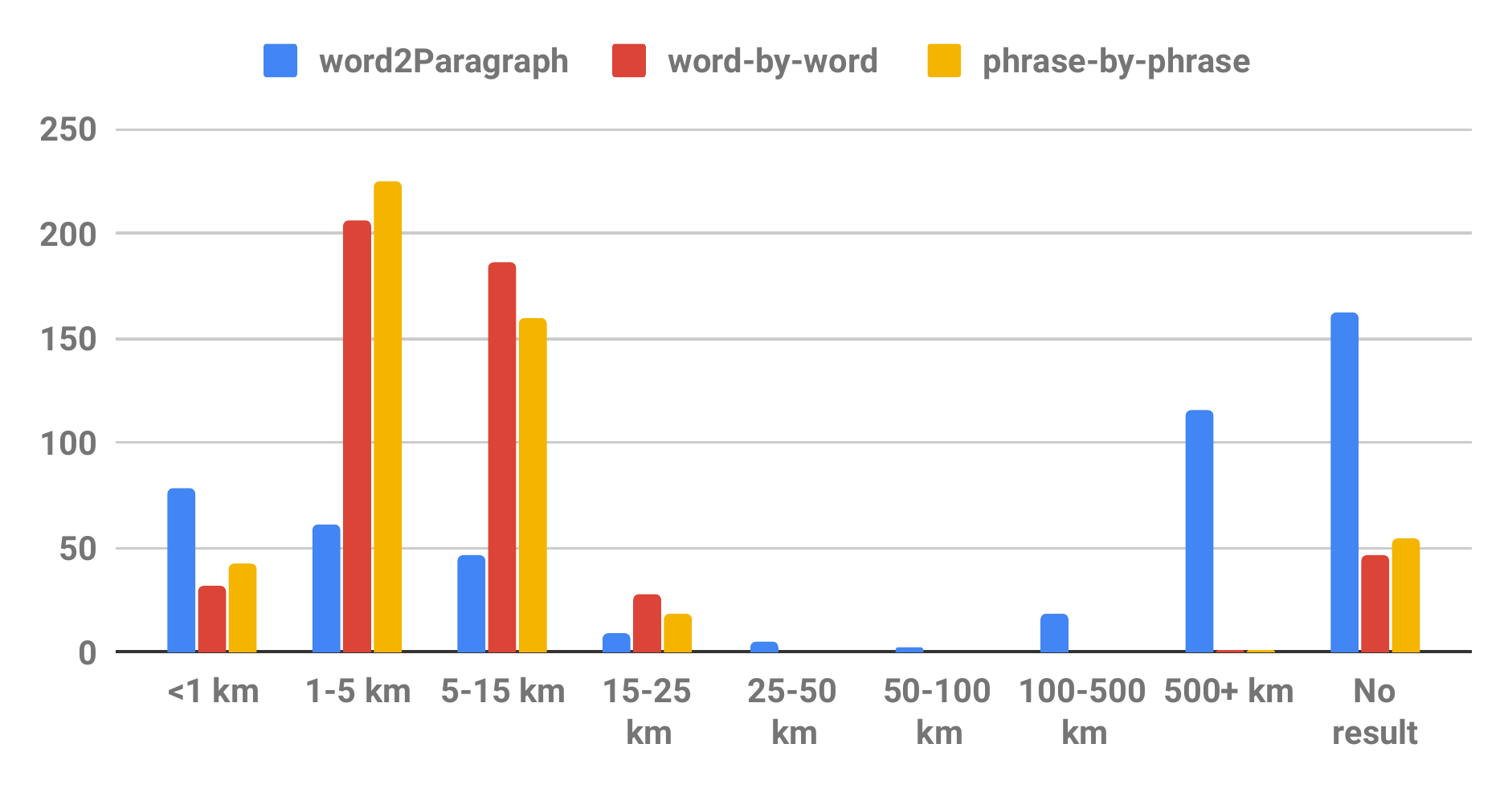}
\end{center}
   \caption{Comparison of the histogram distribution for word2Paragraph ~\cite{tavakkol2019kartta}, word-by-word encoding (ours) and phrase-by-phrase encoding (ours) on the NYPL dataset. Left: word2Paragraph; Middle: word-by-word; Right: phrase-by-phrase. The x-axis denotes the error. It is clear to see that phrase-by-phrase and word-by-word do not have extreme large errors compared with word2Paragraph. }
\label{fig:nypl_hist}
\end{figure}

We also evaluated geolocalization on a much larger dataset -- the \textbf{NYPL} dataset that contains 500 historical map images. This dataset does not have bounding boxes information for individual text regions, so we used Google Vision API for text detection and feed the detection result to the downstream models. Since the number of map images is large, we could not list the result for each map. We instead provide the error distribution histogram in Figure ~\ref{fig:nypl_hist}. The \texttt{word-by-word} and \texttt{phrase-by-phrase} do not have as many large errors as baseline \texttt{word-to-paragraph} method. Also, the number of map images that the model fails to handle got decreased with \texttt{word-by-word} and \texttt{phrase-by-phrase} method.

\begin{table*}
\setlength\tabcolsep{4 pt}
\centering
  \begin{tabular}{c|c|cc|cc|cc|cc}
    \toprule
    & \multirow{2}{*}{Map Name }  & 
      \multicolumn{2}{c}{\textbf{Wrd2Paragraph }} &
      \multicolumn{2}{c}{\textbf{WrdByWrd(Ours)}}  &
       \multicolumn{2}{c}{\textbf{PhrasByPhras(Ours)}}  &
      \multicolumn{2}{c}{\textbf{Ground Truth}}  \\
     & & {Lat.} & {Lng.}  & {Lat.} & {Lng.}& {Lat.} & {Lng.} & {Lat.} & {Lng.}  \\
      \midrule
    \multirow{3}{*}{\shortstack[l]{\textbf{Pred.} \\($^{\circ}$) }} & 60-CA-amboy-e1942 & 38.57    &-121.48 & 33.94& -116.83 & 34.25 &    -116.24 & 34.50    &-115.50 \\
   &  60-CA-amboy-e1943-rv1943 & 38.57 & -121.48&  33.97& -116.78 & 34.24 & -116.18 & 34.50 &    -115.50 \\
   &  60-CA-modoclavabed-e1886 & - & - & 37.99    & -121.80 & 41.16&     -121.54 & 41.50 &    -121.50  \\
    \midrule
    \midrule
    & & km & scale& km & scale& km & scale& km & scale \\
    \midrule
    \multirow{3}{*}{ \shortstack[l]{\textbf{Error} }} & 60-CA-amboy-e1942 &699.89  & 4.85& 137.23& 0.95 &\textbf{ 73.38} & \textbf{0.51}& N/A& N/A\\        
   &  60-CA-amboy-e1943-rv1943 & 699.89 & 4.85&    131.60 & 0.91&\textbf{68.78}& \textbf{0.48 }& N/A & N/A\\
   &  60-CA-modoclavabed-e1886 &N/A & N/A& 391.14 & 2.82 & \textbf{37.95} & \textbf{0.27}& N/A & N/A\\
    \bottomrule
  \end{tabular}
  \caption{Geolocalization result on USGS maps. The upper rows are the prediction results from different methods, and the bottom rows are the errors deviated from ground truth. We use the \textbf{Haversine distance} in kilometers and in map scale as the error metric, thus the lower the better. "-" in the entry means no geolocalization result was returned by the model.}
  \label{tab:usgs_geolocalization_prediction}
\end{table*}

\subsection{Case Study of Linked Map Metadata}\label{sec:case_study}

\begin{figure}[h]
\begin{center}
\includegraphics[width=1.05\linewidth]{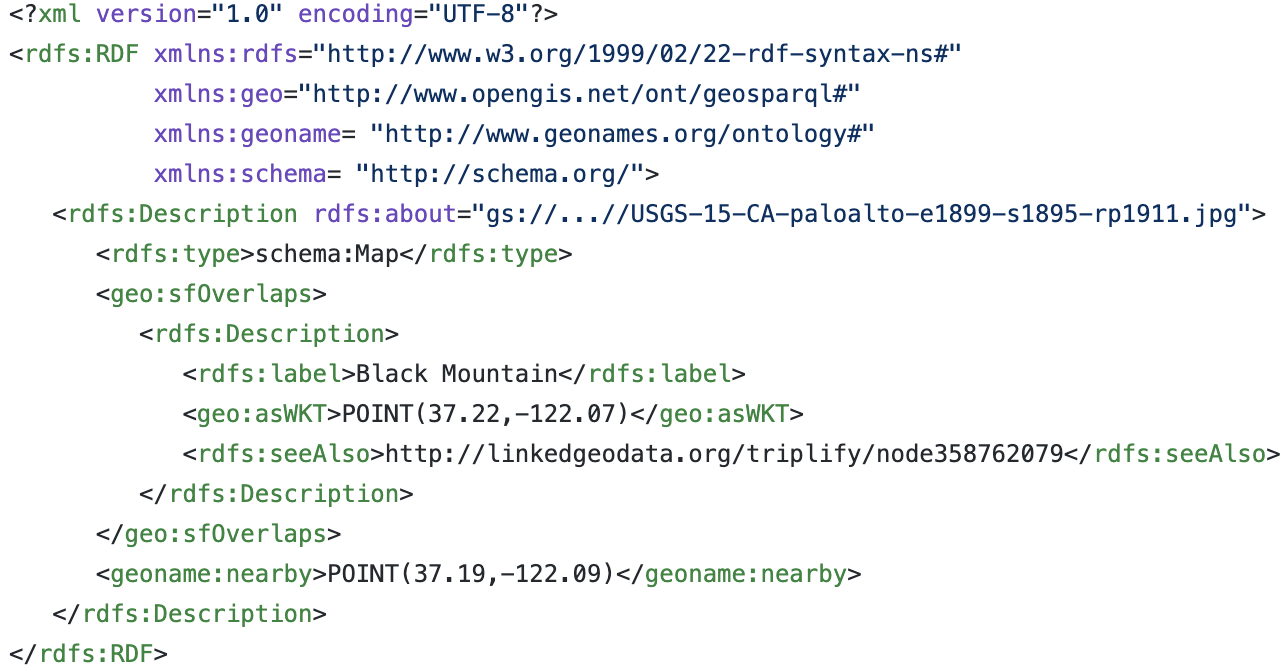}
\end{center}
   \caption{Linked historical map in RDF with XML syntax}
\label{fig:xml}
\end{figure}

In Section ~\ref{sec:entity_matching}, we presented the RDF schema adopted for linked-historical maps, and explained that the structure enables us to jointly query with other external databases. In this section, we provide an example of a spatial query using SPARQL and show the RDF schema graph in XML in Figure ~\ref{fig:xml}.

Based on this, we run a query that finds the historical maps which contains mountain(s) higher than 1km. From GeoLinkedData, the query first select all the mountains and filter out the ones whose elevations are less than one kilometer. Then it checks the \texttt{seeAlso} predicate and backtrack to the text regions that are associated with those mountains. At last, our system returns the names of the maps that overlaps with those text regions. 
\begin{figure}[h]
\begin{center}
\includegraphics[width=0.9\linewidth]{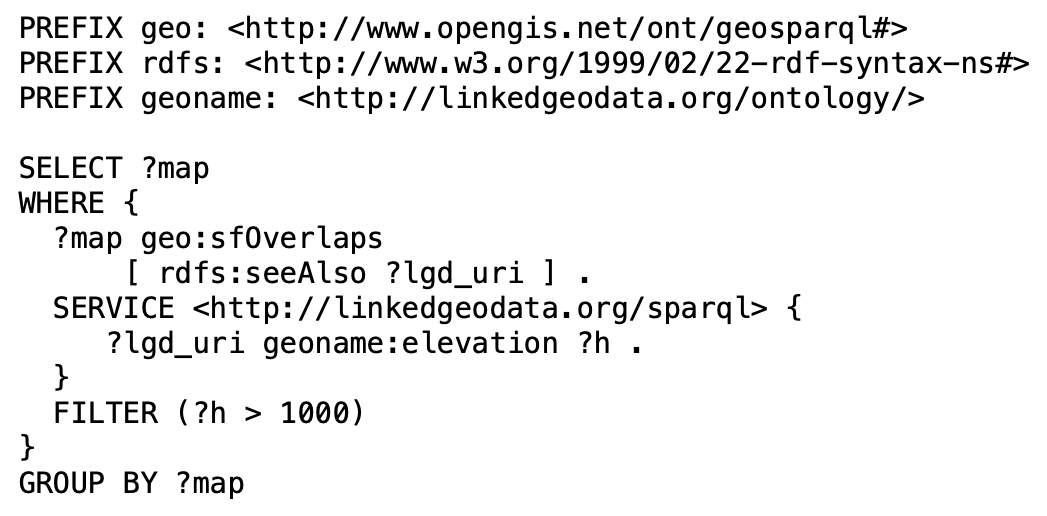}
\end{center}
   \caption{Sample SPARQL query to retrieve map names that contain peaks higher than 1km}
\label{fig:sparql}
\end{figure}


\subsection{Error Analysis}
Since geolocalization and entity linking heavily depend on the result of phrase generation, it is crucial to analyze the errors in the phrase generation step. In Table ~\ref{tab:phrase_prediction}, we have already shown the precision and recall of the generated phrases, and we can dig deeper into that by looking at the cases that caused the error. There are mainly three types of error: 1) pairwise linkage prediction error that happens during the text linking step 2) linkage graph generation error that depends on the method of graph construction 3) element sorting error for deciding the correct sequence of the text regions that belong to the same phrase. The first type of error has been analyzed in previous sections and the second type of error depends on the graph construction. Thus we analyze the third type of error in this section. In Table ~\ref{tab:error_analysis}, we show the precision and recall for both unordered word sets (i.e., elements in connected component) and phrase prediction result. The values in the ``Unordered" column can be seen as the upper bound of the phrase prediction given the current graph construction setting since it ignores the sequence of the word in the location phase, which is the same as the perfect ordering of words. By comparing these two columns, we could see that our simple ordering method, which sorts according to the x-axis coordinate works fine for this type of maps, as the difference between ``Unordered" and "Ordered" is pretty small. These two columns analyzed the third type of error. 

\begin{table}[h]
\centering
\setlength\tabcolsep{4 pt}
  \begin{tabular}{c|ccccc}
    \toprule
    Err (km) & Unordered & Ordered & \# miss & \# add & \#GT \\
    \midrule
    amboy-e1942 & 46.58/65.79 & 44.72/63.16 & 89& 0& 114\\
    amboy-rv1943 & 52.94/69.68 &52.45/69.03 & 95 &4 & 155\\
    modoclavabed & 31.28/51.85 & 31.28/51.85 & 123 &0 & 108\\
    \bottomrule
  \end{tabular}
  \caption{Error analysis for generated phrases. In the table, ``Unordered" column shows the precision and recall when we do not consider the phrase ordering, ``Ordered" means we consider both the words and the ordering for the phrase. \#miss counts the number of errors caused by missing words,  \#add counts the number of errors caused by adding words and \#GT means the number of ground truth phrases contained in the map. }
  \label{tab:error_analysis}
  \vspace{-2em}
\end{table}

\section{Deployment}
Tavakkol et al. ~\cite{tavakkol2019kartta} introduce an open-source project, Kartta Labs, to organize the world’s historical maps and make them universally accessible and useful. They define a modular design for the project to let it thrive on a collaborative community development effort. The current study is done in collaboration with the Kartta Labs team and with the purpose of being deployed in the project. We, specifically, design and implement the Geolocalizer and Linker modules of Kartta Labs. To learn more, see  https://github.com/kartta-labs/Project.

To use our model in production, we plan to deploy it on Google Cloud Functions, which is an event-driven serverless compute platform. Our Cloud Function deployment runs the model as a service and makes it accessible through HTTP requests. In our preliminary implementation, the map images are stored on Google Cloud Storage and each image has been assigned with a URI. The model cloud function takes the URI of the map image as the input. It then sends it to the Google Vision API to retrieve the textual information of the map. The response from the Vision API is preprocessed and sent to the model for generating location phrases. The generated phrases are sent to Google Geocoding API. Finally, the geolocation of the map is determined based on the candidate locations returned from the Geocoding API. The generated phrases, and the geolocation of the map is returned to the client in a JSON response which can be used for downstream tasks such as entity linking.

After successful deployment and productionization, our implementation will replace the word2paragraph geolocalizer in Kartta Labs Warper web application. Warper is a web application that lets the users upload a historical map and georectify it by finding control points on the historical map and corresponding points on a contemporary base map. Once the user uploads a map, the map is sent to the geolocalizer. The base map is loaded according to the estimated geolocation of the historical map. This lets the user immediately start the georectification process without the need to navigate the world map and load the base map in an appropriate location.

\section{Related Work}
\subsection{Text Detection and Text Linking}  
Historical map processing has attracted a lot of attention in recent years\cite{chiang2020using,li2019generating,chiang2016assessing}, but the textual information extraction on maps still remains a challenging problem. Many networks have been developed to detect and recognize text content from scene images such as EAST ~\cite{zhou2017east} and TextBox++ ~\cite{liao2018textboxes++} ; however they do not provide any information about the relation between detected words. Some words can be joined together to form phrases. The task of linking related words can be seen as finding similar features in a hidden embedding space. Deep metric learning aims to explicitly capture the similarity among the features and learn an embedding space where similar features are grouped together, and dissimilar features are quite far away. Siamese network ~\cite{guo2017learning, taigman2014deepface} and Triplet network ~\cite{hoffer2015deep} are the most widely used structures for the deep metric learning models. Siamese network computes pairwise similarity and tries to push the similar pairs closer and dissimilar pairs far away. In Triplet networks, instead of comparing \textit{two} feature points, it compares 3-element tuple. Another work that is closely related to text linking is weakly supervised segmentation ~\cite{khoreva2017simple}, where only partial labels are given. We take advantage of both deep metric learning and weakly supervised segmentation to generate linking results.

\subsection{Geolocalization based on Text Content }
Geolocalization of historical map is estimating the location of the map on Earth. Some work has been done to geolocalize the social network users based on the text content that the user posted ~\cite{ferrari2011extracting, xu2015geolocalized}, but not much has been done to geolocalize map images. The work that is most closely related to ours is ~\cite{tavakkol2019kartta}, which utilized Google NLP API and Google Geocoding API to produce latitude and longitude coordinates. This work takes in all the location-related words detected from the map and concatenates them into paragraphs. The Geocoding API reads the paragraph as input and produces one location. We proposed word-by-word and phrase-by-phrase geocoding that takes words or phrases instead to generate candidate locations, then further clusters the candidate locations to produce the geolocalization output.

\section{Discussion and Future Work}
This paper presented an end-to-end approach to generate rich, linked metadata from historical maps. In addition to the complete approach that addresses the real-world problem of finding and indexing historical map images, our contribution is two-fold: First, we proposed a model to harness both textual information and visual information to generate location phrases. Second, we designed a method to geolocalize a map image using the location phrases generated from the first step. The proposed method constitutes an important building block for efficient, large-scale information extraction processing chains from historical map archives. We plan to investigate sophisticated ontologies to represent the metadata so that the metadata can support more types of queries and record important information, such as the provenance information. We also plan to expand the phrase generation module to automatically identify the source (e.g., map publisher) and temporal information from a map image.

\section*{Acknowledgements}
This material is based upon work supported in part by the National Science Foundation under Grant Nos. IIS 1564164 (to the University of
Southern California) and IIS 1563933 (to the University of Colorado at Boulder), NVIDIA Corporation, and the USC Undergraduate Research Associates Program.
\bibliographystyle{ACM-Reference-Format}
\bibliography{reference}

\end{document}